\documentclass[runningheads]{llncs}
\usepackage[T1]{fontenc}
\usepackage{graphicx}
\usepackage{amsmath}
\usepackage{booktabs}
\usepackage{hyperref}
\usepackage{cleveref}
\usepackage{float}
\usepackage{appendix}
\usepackage{rotating}

\usepackage{caption}
\usepackage{subcaption}

\usepackage{array}
\newcolumntype{L}[1]{>{\raggedright\let\newline\\\arraybackslash\hspace{0pt}}m{#1}}
\newcolumntype{C}[1]{>{\centering\let\newline\\\arraybackslash\hspace{0pt}}m{#1}}
\newcolumntype{R}[1]{>{\raggedleft\let\newline\\\arraybackslash\hspace{0pt}}m{#1}}

\usepackage{color}

\newcommand{\connectfour}{\textsc{ConnectFour}}

\begin{document}
\title{Strongly Solving $7 \times 6$ \connectfour{} on Consumer Grade Hardware}
%
%
\author{Markus Böck}
\authorrunning{M. Böck}
%
\institute{TU Wien, Vienna, Austria\\
\email{markus.h.boeck@tuwien.ac.at}}
\maketitle              
\begin{abstract}
While the game \connectfour{} has been solved mathematically and the best move can be effectively computed with search based methods, a strong solution in the form of a look-up table was believed to be infeasible.
In this paper, we revisit a symbolic search method based on binary decision diagrams to produce strong solutions.
With our efficient implementation we were able to produce a 89.6 GB large look-up table in 47 hours on a single CPU core with 128 GB main memory for the standard $7 \times 6$ board size.
In addition to this win-draw-loss evaluation, we include an alpha-beta search in our open source artifact to find the move which achieves the fastest win or slowest loss.

\end{abstract}
\section{Introduction}\label{s1}

\connectfour{} is a game in which two players take turns placing colored discs into a two-dimensional board.
A players win if they manage to form a horizontal, vertical, or diagonal line of four of their own disks.
The game is a draw if the board is fully filled and no such lines were achieved.
Piece placement is restricted as players are only allowed to place their disk in the lowest available cell within the column.
In the real world, this restriction is enforced by suspending the board vertically such that the disk fall straight down.
The most commonly-used board size is 7 columns $\times$ 6 rows.

In 1988 Allen~\cite{allen2010complete} was the first to solve \connectfour{} by describing winning strategies.
Published only 15 days later, Allis~\cite{allis1988knowledge} independently came up with a knowledge-based approach to solve the game.
With the advancements of computer hardware, explicit search-based approaches alá alpha-beta pruning~\cite{knuth1975alphabeta} were used by Tromp~\cite{tromp-8-ply} to build a database which stores the win-draw-loss evaluation for each 8-ply position.
This effectively strongly solved $7 \times 6$ \connectfour{} as solving positions deeper than eight plies can be done in seconds with explicit search.
Later, Tromp~\cite{tromp2008solving} weakly solved \connectfour{} instances with $\text{width} + \text{height} < 16$.
Recently, Steininger~\cite{Steininger2024} extended these solutions for board sizes where $\text{width} + \text{height} = 16$.

While explicit search traverses one state at a time, so called symbolic search methods handle sets of states at once.
Edelkamp and Kissmann~\cite{edelkamp2011complexitybdd} pioneered the use of binary decision diagrams (BDDs) to encode sets of states where a single state is represented by a conjunction of binary variables.
Kissmann and Edelkamp~\cite{kissmann2010layer} developed a layered approach in which all game positions at a given ply are encoded by a BDD.
Thus, at the end of the so-called forward pass we have 42 disjoint BDDs encoding all $7 \times 6$ \connectfour{} positions.
The approach then proceeds by performing a retrograde analysis starting at all fully filled positions at ply 42, splitting them into win-, draw-, and lost-BDDs, and propagating the classification back ply-by-ply to the empty root position.

The algorithmic complexity of BDDs depends heavily on the state encoding.
Edelkamp and Kissmann~\cite{edelkamp2011complexitybdd} proved the unfortunate fact that for a natural encoding the number of nodes required to encode all positions scales exponentially in the number of columns or rows.
As a consequence, they were only able to solve the $6 \times 6$ board configuration with 64 GB RAM in 22:38 hours.
To the best of our knowledge Edelkamp et al.~\cite{edelkamp2014symbolicexplicit} constitutes their last published effort to solve the standard $7 \times 6$ board configuration.
Their hybrid between explicit and symbolic search trades of memory with computational requirements was estimated to produce a solution in 93 days on a 192 GB machine.

In this paper, we revisit the approach of Kissmann and Edelkamp~\cite{kissmann2010layer}.
With a different state encoding and minor algorithmic improvements, we were able to solve the $7 \times 6$ board configuration in 47 hours on a single core of the AMD Ryzen 5950x CPU with 128 GB main memory.
By writing the solution as BDDs to disk, they take only 89.6 GB storage.
We were also able to reproduce the counting of unique positions of Edelkamp and Kissmann~\cite{edelkamp2011complexitybdd} in 2:15 hours and the solution of the $6 \times 6$ board configuration in 2:13 hours, both using 32 GB RAM.
Lastly, we implemented a rudimentary alpha-beta search on top of our win-draw-loss solution to find the move which achieves the fastest win or slowest loss.
The source code to reproduce our results and to query the win-draw-loss table is openly available at \textcolor{blue}{\url{https://github.com/markus7800/Connect4-Strong-Solver}}.

\section{Background}

\subsection{Binary Decision Diagrams}

Binary Decision Diagrams (BDDs) are a data structure to represent Boolean functions over variables $x_i$ as a rooted, directed, acyclic graph~\cite{andersen1997introductionbdd,bryant1986graph}.
Each BDD contains two terminal nodes labeled 0 (\texttt{false}) and 1 (\texttt{true}).
Each internal node is a decision node associated with one Boolean variable $x_i$ and two child nodes called low and high child.
The edges to the low and high child represent the assignment of the variable $x_i$ to \texttt{false} and \texttt{true}, respectively.
Thus, a path from the root node to the 1 terminal node, represents a possibly partial variable assignment for which the encoded function is \texttt{true}.
Likewise, a path to the 0 terminal corresponds to a \texttt{false}-assignment.
Importantly, BDDs allow the direct application of Boolean operations.
If the ordering of variables is the same for each path from root to terminal, then the BDD is called ordered.
If the graph is unique up to isomorphism, the BDD is called reduced.
In this work, we work with these reduced ordered binary decision diagrams (ROBDDs).

\subsection{BDDs for Symbolic Search for Two-Player Games}\label{sec:symbolic-search}

In symbolic search, we represent \emph{sets of states} as Boolean functions $f$ over some set of variables $S$ ~\cite{edelkamp2014symbolicexplicit,kissmann2010layer}.
We have $f(S) = \texttt{true}$ if the state corresponding to the variable assignments $S$ belongs to the state of sets $f$, otherwise  $f(S) = \texttt{false}$.
The way we encode a \connectfour{} state in terms of Boolean variables is described in \Cref{sec:state-encoding}, but not important for this section.
Computationally, the Boolean functions are represented as BDDs and for which we can directly apply operations like $\land, \lor, \neg,$ etc.

To encode transitions from one set of states to another following the game rules, we need a copy of the variables $S'$ and a transition relation $\text{trans}(S,S')$.
It typically has the form $\text{trans}(S,S') = \bigvee_a \text{pre}_a(S) \land \text{eff}_a(S') \land \text{frame}_a(S,S')$, where for each game action $a$ we have a pre-condition, effect, and a frame that determines the variables that remain unchanged when applying $a$.
With the transition relation, we can define the image and pre-image operations:
\begin{align*}
\text{image}(f) = \exists S\colon f(S) \land \text{trans}(S,S'), \quad
\text{pre-image}(f) = \exists S'\colon \text{trans}(S,S') \land f(S')
\end{align*}
To solve a game with win/draw/loss outcome, we first perform a \emph{forward pass}, where we compute $\text{states}_i$ -- the set of all positions at a given ply $i$.
We start at the initial game state $\text{states}_0 = \text{initial\_state}$ and compute recursively
\begin{align*}
\text{states}_{i+1} \gets \text{image}(\text{states}_i \land \neg\text{terminal})[S' \to S],
\end{align*}
where $\text{terminal}$ is the Boolean function representing the set of all terminal states.
To keep representing the states in terms of variables $S$, we have to replace the variables $S'$ with $S$ after performing the $\text{image}$ operation, denoted by $[S' \to S]$.

After computing the set of states at each ply, we propagate the game outcome backwards with following rules
\begin{align*}
\text{win}_{i} &\gets \text{pre-image}(\text{lost}_{i+1}[S \to S']) \land (\text{states}_i \land \neg \text{terminal}),\\
\text{draw}_{i} &\gets \text{pre-image}(\text{draw}_{i+1}[S \to S']) \land (\text{states}_i \land \neg \text{terminal} \land \neg \text{win}_{i} ),\\
\text{lost}_{i} &\gets (\text{states}_i \land \neg \text{terminal} \land \neg \text{win}_{i} \land \neg \text{draw}_{i}) \lor (\text{states}_i \land \text{terminal}).
\end{align*}
Here, we leverage the fact that a win for one player is a loss for the other.
These Boolean functions are from the perspective of the player to move.
Thus, at the maximum ply $N$, the current player cannot move and thus we have a draw on non-terminal and a loss on terminal states and initialise
$\text{win}_{N} = \text{false}$, $\text{draw}_{N} = \text{states}_N \land \neg\text{terminal}$, and $\text{lost}_{N} = \text{states}_N \land \text{terminal}$.

\section{Implementation Details}

\subsection{State Encoding}\label{sec:state-encoding}
\begin{figure}[H]
     \centering
     \begin{subfigure}[b]{0.49\textwidth}
         \centering
         \includegraphics[width=\textwidth]{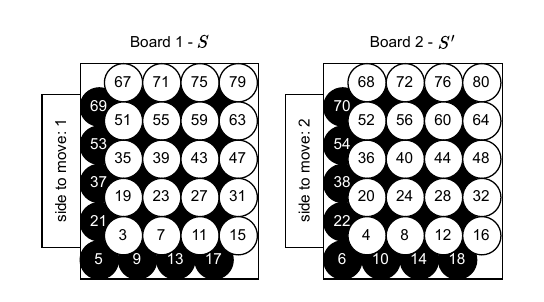}
         \caption{Standard row-wise encoding:\\first-player white, second-player black}
         \label{fig:standard-enc}
     \end{subfigure}
     \hfill
     \begin{subfigure}[b]{0.49\textwidth}
         \centering
         \includegraphics[width=\textwidth]{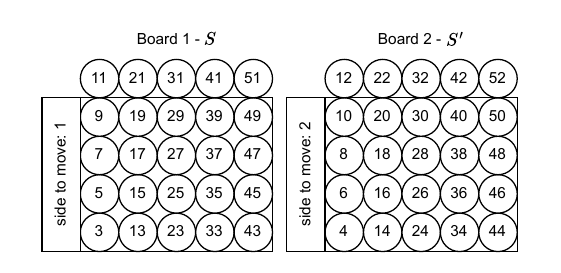}
         \caption{Compressed column-wise encoding:\\an additional row is needed}
         \label{fig:comp-enc}
     \end{subfigure}
     \hfill
     \caption{Variable ordering for different ways of encoding \connectfour{} boards. $4 \times 5$ board on the left, $5\times 4$ board on the right.}
\end{figure}

There are multiple ways to encode a \connectfour{} position with Boolean variables.
Edelkamp and Kissmann~\cite{edelkamp2011complexitybdd} use $2\cdot(\text{width}\cdot\text{height}) + 1$ variables: one variable denotes the side-to-move and there are two variables per board cell indicating whether it is empty, occupied by the first player, or occupied by the second player.
For this cell-wise encoding, Edelkamp and Kissmann~\cite{edelkamp2011complexitybdd} proved that the algorithmic complexity of counting all unique positions is exponential in $\min(\text{width},\text{height})$, independent of variable ordering.
We have implemented their encoding for both a row-wise and column-wise order.
While they have not explicitly mentioned how to order the variables belonging to the same cell we take following approach: player-1-board-1, player-1-board-2, player-2-board-1, player-2-board-2, as illustrated in \Cref{fig:standard-enc}.
Here board-1 corresponds to the variable set $S$ and board 2 to the variable set $S'$ as explained in \Cref{sec:symbolic-search}.
\vspace{-1cm}
\vspace{-0.5cm}
Inspired by bitboard representations~\cite{tromp2008solving,Herzberg}, we have also implemented a more compressed encoding with only $\text{width} \cdot (\text{height} + 1) + 1$ variables, illustrated in \Cref{fig:comp-enc}.
In this encoding, each cell has only one variable.
The lowest available cell per column is always \texttt{true}.
All cells below the lowest available cell are occupied and \texttt{true} if the disc belongs to the first player else \texttt{false} (the disc belongs to the second player).
To facilitate this logic, we need an additional row.
This encoding outperforms the standard encoding for the $7 \times 6$ board configuration, but performs poorly if the number of rows is large.

\subsection{BDD Implementation}
We have implemented a minimal BDD library in C which allows us to implement the computations described in \Cref{sec:symbolic-search}.
No special improvements in the implementation of the primitive BDD operations were made.
The garbage collection is typically a challenging aspect of BDD libraries.
In our implementation, the user specifies the number of nodes desired and the library pre-allocates all nodes.
With a form of reference counting, the user has to manually free nodes to make them available for reuse.
This makes our library fast with a lot of control over memory consumption, but also a bit tedious to work with.

The termination criteria in \connectfour{} is a conjunction of many Boolean formula each representing the condition that there are four discs of the same color in a row (horizontally, vertically, or diagonally) for a different set of cells.
A lot of nodes would be necessary to encode the $\text{terminal}$ BDD as a whole.
Instead, in the implementation of $\text{states}_i \land \neg \text{terminal}$, we iteratively subtract the individual termination condition.
We handle $\text{states}_i \land \text{terminal}$ in a similar fashion.

Lastly, we avoid renaming variables, $[S \to S']$, by computing the mirrored transition relation $\text{trans}'(S',S)$ which switches the roles of variable sets $S$ and $S'$.
We also implemented a special sat-counting routine to count the number of states encoded in a BDD with respect to only $S$ \emph{or} $S'$ instead of $S \cup S'$.

\subsection{Alpha-Beta Search}

In addition to the win-draw-loss evaluation of a position, we like to also find the move that gives the fastest win or prolongs the game as long as possible in the case of a loss.
To this end, we implemented a rudimentary alpha-beta principal variation search~\cite{knuth1975alphabeta}.
It uses the the usual bitboard representation to facilitate fast move and position generation~\cite{ponsblog,tromp2008solving,Herzberg}.
We leverage the vertical symmetry of the position evaluation in a transposition table of size $2^{28}$ that always overwrites old entries in case of collision.
Moves are sorted by their distance to the central board column and by the number of threats they generate.
In this context, a threat is an alignment of three discs on the board waiting to be completed to connect four discs and win the game.
We prune some positions where we can statically detect a win, draw, or loss.
For example, a player cannot prevent two threats at the same time, thus losing the game.
Lastly, in the lower depths, we probe the win-draw-loss table to avoid searching losing moves in a won position.
For a state-of-the-art \connectfour{} search engine see Steininger~\cite{Steininger2024}.






\section{Results}
\subsection{Counting Unique Positions}
Before presenting the \connectfour{} solution, we briefly discuss the number of unique game positions computed by the forward pass.  

\begin{figure}[H]
     \centering
     \begin{subfigure}[b]{0.49\textwidth}
         \centering
         \includegraphics[width=\textwidth]{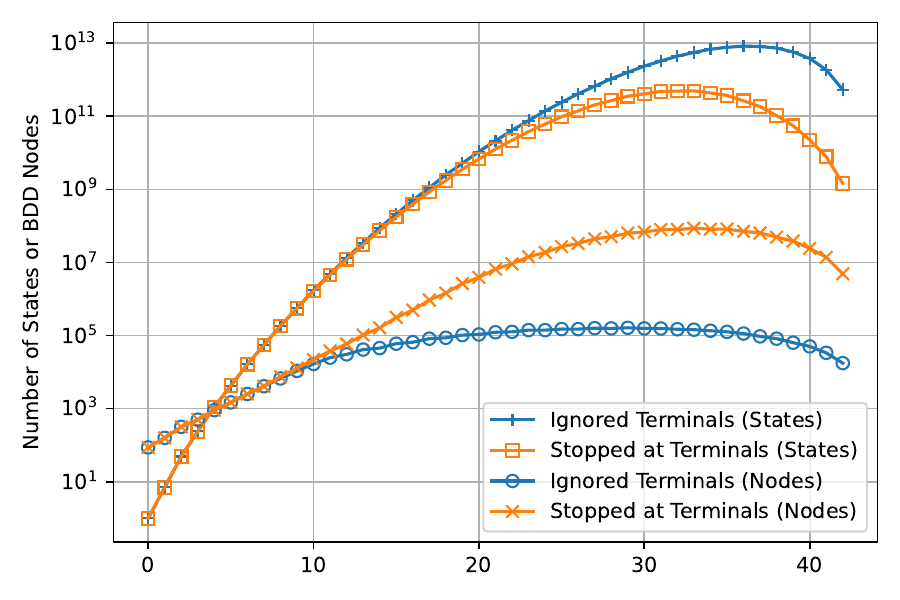}
         \caption{Number of states and BDD nodes at each ply for $7 \times 6$ \connectfour{} with and without termination criterion}
         \label{fig:term-vs-non-term}
     \end{subfigure}
     \hfill
     \begin{subfigure}[b]{0.49\textwidth}
         \centering
         \includegraphics[width=\textwidth]{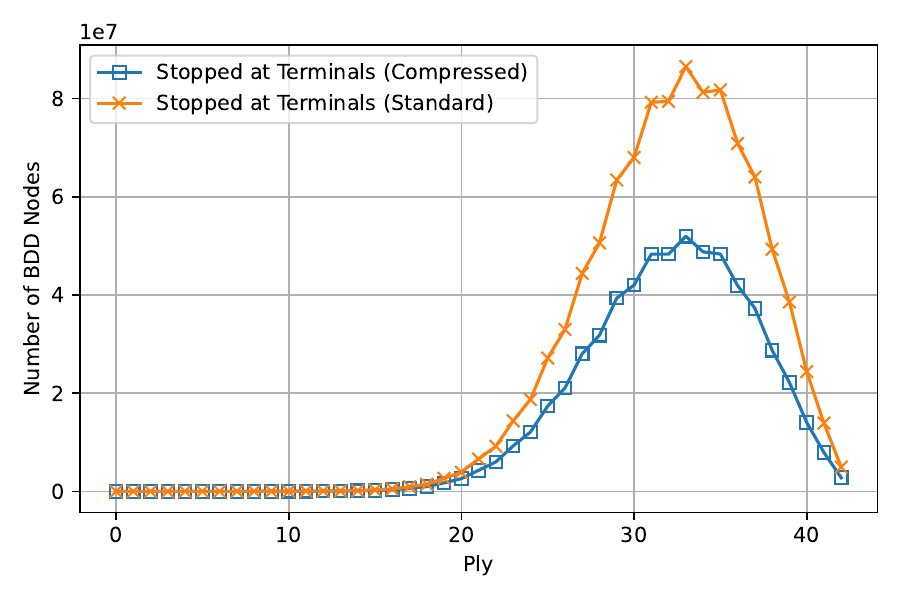}
         \caption{BDD nodes required to represent all states at each ply: compressed versus standard state encoding.}
         \label{fig:compenc-vs-standard}
     \end{subfigure}
     \hfill
     \caption{Investigation of BDD sizes for the $7 \times 6$ board configuration.}
\end{figure}

In \Cref{fig:term-vs-non-term}, you can see the number of positions and the size of the BDD to encode them at each ply for the $7 \times 6$ board computed in 3:34 hours with 32 GB RAM.
In addition, you can see the same quantities if we remove the termination condition.
This graph effectively reproduces the results presented in Edelkamp and Kissmann~\cite{edelkamp2011complexitybdd}.
Presumably due to slightly different variable ordering, we find that it takes 95,124,612 nodes to encode all positions into a single BDD instead of the 84,088,763 reported Edelkamp and Kissmann~\cite{edelkamp2008symbolicclass}.
As can be seen in \Cref{fig:compenc-vs-standard}, with the compressed encoding we can lower this number to 59,853,336 and computation takes only 2:15 hours.
With our implementations, we can confirm all unique positions counts of Tromp~\cite{tromplayground} and produce novel counts included in the Appendix.
For example, it took almost two days to compute that the $7 \times 7$ game has 161,965,120,344,045 unique positions.

\subsection{Solving \connectfour{}}

With our implementation and the compressed encoding, we were able to confirm the solution for the $6 \times 6$ game Edelkamp et al.~\cite{edelkamp2014symbolicexplicit} in 2:13 hours with 32 GB RAM.
We were able to produce the 89.6 GB win-draw-table for the $7 \times 6$ game in 47 hours with 128 GB RAM.
The memory was not sufficient for solving with the standard encoding.
In the Appendix, we list the number of won, drawn, lost, and terminal position at each ply.
This table agrees with the partial solution presented in Edelkamp et al.~\cite{edelkamp2014symbolicexplicit}.
With over 1.1 billion nodes and 10.2 GB, ply 27 takes the most storage to store.
However, since $\text{draw} = \neg \text{lost} \land \neg \text{win}$, we do not need to store the draw BDD reducing the amount of storage required significantly.

By leveraging the win-draw-loss evaluation in the alpha-beta search, we can confirm in 9.2 seconds that the first player wins in 41 plies by playing in the center.
Not using the table increases the number of explored positions by a factor of 6 and takes 128 seconds.
To facilitate almost instantaneous position evaluation, we generated an openingbook by evaluating all 184,275 8-ply positions with the alpha-beta search making use of the win-draw-loss.
Using all 16 cores, this took 5:11 hours.
The full win-draw-loss solution and the openingbook can be found in Böck~\cite{boeck_2024_14582823}.

\section{Conclusion}
In this work, we revisited symbolic search for solving $7 \times 6$ \connectfour{}.
Due to exponential algorithmic complexity, it was believed that this method cannot be scaled to this board configuration.
With our open-source implementation, we demonstrate that, in fact, this instance can be solved with moderate hardware requirements.
By storing the solution directly as binary decision diagrams, we are the first to produce a look-up-table-like solution.
As a by-product, we effectively reproduced prior research concerned with counting unique game positions and present novel counts including the $7 \times 7$ board.  

\newpage
%
%
%
\bibliographystyle{splncs04}
\bibliography{references}
\newpage

\section*{Appendix}
\vspace{-1cm}

\begin{table}[H]
\centering
\caption{Number of positions for several $w \times h$ board configurations with computation time, percentage spent garbage collecting, RAM used, and maximum number of BDD nodes allocated at the same time.}
\begin{tabular}{R{5mm}R{5mm}R{35mm}R{20mm}R{13mm}R{15mm}R{20mm}}
\toprule
$w$ & $h$ & \# positions & time & GC & RAM & \#nodes alloc. \\
\midrule
1 & 1 & 2 & 0:00:01.06 & 54.41\% & 3.42GB & 61 \\
1 & 2 & 3 & 0:00:02.35 & 54.24\% & 3.42GB & 222 \\
1 & 3 & 4 & 0:00:03.25 & 48.33\% & 3.42GB & 468 \\
1 & 4 & 5 & 0:00:02.99 & 42.06\% & 3.42GB & 825 \\
1 & 5 & 6 & 0:00:03.12 & 40.28\% & 3.42GB & 1,306 \\
1 & 6 & 7 & 0:00:03.17 & 43.47\% & 3.42GB & 1,923 \\
1 & 7 & 8 & 0:00:03.87 & 38.74\% & 3.42GB & 2,688 \\
1 & 8 & 9 & 0:00:04.21 & 35.62\% & 3.42GB & 3,613 \\
1 & 9 & 10 & 0:00:06.06 & 51.84\% & 3.42GB & 4,710 \\
1 & 10 & 11 & 0:00:08.95 & 58.61\% & 3.42GB & 5,991 \\
1 & 11 & 12 & 0:00:09.10 & 60.28\% & 3.42GB & 7,468 \\
1 & 12 & 13 & 0:00:10.46 & 66.53\% & 3.42GB & 9,153 \\
1 & 13 & 14 & 0:00:13.70 & 71.95\% & 3.42GB & 11,058 \\
2 & 1 & 5 & 0:00:01.93 & 58.09\% & 3.42GB & 203 \\
2 & 2 & 18 & 0:00:02.86 & 46.32\% & 3.42GB & 800 \\
2 & 3 & 58 & 0:00:03.68 & 38.29\% & 3.42GB & 1,957 \\
2 & 4 & 179 & 0:00:04.29 & 33.52\% & 3.42GB & 3,777 \\
2 & 5 & 537 & 0:00:08.48 & 54.39\% & 3.42GB & 6,330 \\
2 & 6 & 1,571 & 0:00:10.42 & 66.42\% & 3.42GB & 9,700 \\
2 & 7 & 4,587 & 0:00:14.47 & 69.97\% & 3.42GB & 13,971 \\
2 & 8 & 13,343 & 0:00:13.47 & 71.62\% & 3.42GB & 19,227 \\
2 & 9 & 38,943 & 0:00:23.26 & 76.53\% & 3.42GB & 25,552 \\
2 & 10 & 113,835 & 0:00:19.88 & 75.92\% & 3.42GB & 33,030 \\
2 & 11 & 333,745 & 0:00:26.56 & 76.63\% & 3.42GB & 41,745 \\
2 & 12 & 980,684 & 0:00:22.12 & 76.01\% & 3.42GB & 51,781 \\
2 & 13 & 2,888,780 & 0:00:20.44 & 78.30\% & 3.42GB & 63,222 \\
3 & 1 & 13 & 0:00:03.35 & 46.73\% & 3.42GB & 426 \\
3 & 2 & 116 & 0:00:03.08 & 39.53\% & 3.42GB & 1,836 \\
3 & 3 & 869 & 0:00:05.32 & 40.88\% & 3.42GB & 4,567 \\
3 & 4 & 6,000 & 0:00:11.00 & 67.31\% & 3.42GB & 10,781 \\
3 & 5 & 38,310 & 0:00:19.85 & 70.88\% & 3.42GB & 18,520 \\
3 & 6 & 235,781 & 0:00:23.68 & 76.04\% & 3.42GB & 29,974 \\
3 & 7 & 1,417,322 & 0:00:20.59 & 74.19\% & 3.42GB & 75,162 \\
3 & 8 & 8,424,616 & 0:00:28.97 & 72.81\% & 3.42GB & 172,756 \\
3 & 9 & 49,867,996 & 0:00:26.03 & 59.16\% & 3.42GB & 422,983 \\
3 & 10 & 294,664,010 & 0:00:35.54 & 40.98\% & 3.42GB & 1,010,556 \\
3 & 11 & 1,741,288,730 & 0:01:10.14 & 28.39\% & 3.42GB & 2,381,296 \\
3 & 12 & 10,300,852,227 & 0:04:03.01 & 25.93\% & 3.42GB & 5,427,670 \\
3 & 13 & 61,028,884,959 & 0:09:06.17 & 18.47\% & 3.42GB & 13,049,070 \\
4 & 1 & 35 & 0:00:03.03 & 46.03\% & 3.42GB & 745 \\
4 & 2 & 741 & 0:00:04.17 & 33.98\% & 3.42GB & 3,426 \\
4 & 3 & 12,031 & 0:00:10.67 & 66.98\% & 3.42GB & 8,853 \\
4 & 4 & 161,029 & 0:00:27.73 & 85.17\% & 3.42GB & 17,799 \\
4 & 5 & 1,706,255 & 0:01:01.09 & 86.01\% & 3.42GB & 129,589 \\
4 & 6 & 15,835,683 & 0:01:07.39 & 74.44\% & 3.42GB & 561,346 \\
4 & 7 & 135,385,909 & 0:01:18.26 & 44.51\% & 3.42GB & 1,648,551 \\
4 & 8 & 1,104,642,469 & 0:03:01.01 & 27.01\% & 3.42GB & 4,307,162 \\
4 & 9 & 8,754,703,921 & 0:09:25.69 & 20.52\% & 3.42GB & 11,478,742 \\
4 & 10 & 67,916,896,758 & 0:19:22.84 & 16.40\% & 13.69GB & 29,923,914 \\
4 & 11 & 519,325,538,608 & 1:06:08.86 & 12.89\% & 13.69GB & 80,053,135 \\
4 & 12 & 3,928,940,117,357 & 3:36:00.93 & 9.23\% & 13.69GB & 220,438,835 \\
4 & 13 & 29,499,214,177,403 & 9:42:44.88 & 9.78\% & 54.76GB & 608,399,867 \\
\bottomrule
\end{tabular}

\end{table}
\newpage
\begin{table}[H]
\centering
\begin{tabular}{R{5mm}R{5mm}R{35mm}R{20mm}R{13mm}R{15mm}R{20mm}}
\toprule
$w$ & $h$ & \# positions & time & GC & RAM & \#nodes alloc.\\
\midrule
5 & 1 & 96 & 0:00:03.32 & 43.38\% & 3.42GB & 1,172 \\
5 & 2 & 4,688 & 0:00:06.98 & 58.01\% & 3.42GB & 5,666 \\
5 & 3 & 158,911 & 0:00:12.97 & 73.18\% & 3.42GB & 15,062 \\
5 & 4 & 3,945,711 & 0:01:00.14 & 86.09\% & 3.42GB & 129,415 \\
5 & 5 & 69,763,700 & 0:01:21.73 & 64.22\% & 3.42GB & 1,230,084 \\
5 & 6 & 1,044,334,437 & 0:04:52.75 & 21.07\% & 3.42GB & 8,235,134 \\
5 & 7 & 14,171,315,454 & 0:25:17.41 & 15.38\% & 3.42GB & 25,600,589 \\
5 & 8 & 182,795,971,462 & 0:41:32.27 & 14.91\% & 27.38GB & 69,165,125 \\
5 & 9 & 2,284,654,770,108 & 2:32:23.19 & 11.96\% & 54.76GB & 201,469,478 \\
6 & 1 & 267 & 0:00:03.00 & 42.70\% & 3.42GB & 1,719 \\
6 & 2 & 29,737 & 0:00:10.52 & 66.09\% & 3.42GB & 8,652 \\
6 & 3 & 2,087,325 & 0:00:24.93 & 75.60\% & 3.42GB & 32,107 \\
6 & 4 & 94,910,577 & 0:01:10.13 & 73.27\% & 3.42GB & 706,477 \\
6 & 5 & 2,818,972,642 & 0:04:11.02 & 23.74\% & 3.42GB & 7,852,100 \\
6 & 6 & 69,173,028,785 & 0:23:07.92 & 14.52\% & 6.85GB & 61,634,539 \\
6 & 7 & 1,523,281,696,228 & 4:58:47.33 & 9.16\% & 27.38GB & 365,099,251 \\
6 & 8 & 31,936,554,362,084 & 17:39:00.27 & 9.35\% & 109.52GB & 1,023,026,899 \\
7 & 1 & 750 & 0:00:03.15 & 38.79\% & 3.42GB & 2,398 \\
7 & 2 & 189,648 & 0:00:15.09 & 71.34\% & 3.42GB & 12,480 \\
7 & 3 & 27,441,956 & 0:00:20.13 & 71.29\% & 3.42GB & 97,803 \\
7 & 4 & 2,265,792,710 & 0:01:54.27 & 45.32\% & 3.42GB & 2,418,545 \\
7 & 5 & 112,829,665,923 & 0:28:21.81 & 16.11\% & 3.42GB & 27,594,037 \\
7 & 6 & 4,531,985,219,092 & 3:22:58.57 & 13.07\% & 27.38GB & 238,538,432 \\
7 & 7 & 161,965,120,344,045 & 44:41:45.37 & 10.20\% & 109.52GB & 1,797,440,896 \\
8 & 1 & 2,118 & 0:00:04.13 & 35.78\% & 3.42GB & 3,221 \\
8 & 2 & 1,216,721 & 0:00:14.79 & 74.06\% & 3.42GB & 17,246 \\
8 & 3 & 362,940,958 & 0:00:19.49 & 64.98\% & 3.42GB & 238,318 \\
8 & 4 & 54,233,186,631 & 0:04:51.88 & 27.88\% & 3.42GB & 5,820,061 \\
8 & 5 & 4,499,431,376,127 & 0:53:32.50 & 18.26\% & 27.38GB & 70,093,438 \\
8 & 6 & 290,433,534,225,566 & 16:00:43.57 & 11.45\% & 54.76GB & 625,763,115 \\
9 & 1 & 6,010 & 0:00:06.07 & 47.99\% & 3.42GB & 4,200 \\
9 & 2 & 7,844,298 & 0:00:19.03 & 72.86\% & 3.42GB & 23,046 \\
9 & 3 & 4,816,325,017 & 0:00:28.97 & 56.19\% & 3.42GB & 478,748 \\
9 & 4 & 1,295,362,125,552 & 0:09:11.61 & 29.47\% & 13.69GB & 11,398,845 \\
9 & 5 & 178,942,601,291,926 & 2:37:32.12 & 17.09\% & 54.76GB & 163,165,045 \\
10 & 1 & 17,120 & 0:00:07.15 & 57.45\% & 3.42GB & 5,347 \\
10 & 2 & 50,780,523 & 0:00:31.32 & 76.73\% & 3.42GB & 36,276 \\
10 & 3 & 64,137,689,503 & 0:00:26.48 & 35.96\% & 3.42GB & 840,337 \\
10 & 4 & 30,932,968,221,097 & 0:20:38.81 & 23.53\% & 13.69GB & 24,226,909 \\
11 & 1 & 48,930 & 0:00:09.47 & 66.06\% & 3.42GB & 6,674 \\
11 & 2 & 329,842,064 & 0:00:24.87 & 75.23\% & 3.42GB & 57,846 \\
11 & 3 & 856,653,299,180 & 0:00:42.25 & 26.89\% & 3.42GB & 1,615,299 \\
11 & 4 & 738,548,749,700,312 & 0:44:45.79 & 21.18\% & 13.69GB & 48,646,573 \\
12 & 1 & 140,243 & 0:00:11.16 & 67.08\% & 3.42GB & 8,193 \\
12 & 2 & 2,148,495,091 & 0:00:23.60 & 73.27\% & 3.42GB & 92,737 \\
12 & 3 & 11,470,572,730,124 & 0:01:58.76 & 32.64\% & 3.42GB & 2,910,928 \\
12 & 4 & 17,631,656,694,578,591 & 1:25:25.41 & 18.07\% & 13.69GB & 88,481,565 \\
13 & 1 & 402,956 & 0:00:12.29 & 68.79\% & 3.42GB & 9,916 \\
13 & 2 & 14,027,829,516 & 0:00:24.42 & 72.20\% & 3.42GB & 147,712 \\
13 & 3 & 153,906,772,806,519 & 0:02:57.83 & 26.38\% & 3.42GB & 4,927,508 \\
13 & 4 & 420,788,402,285,901,831 & 2:03:51.22 & 19.22\% & 27.38GB & 150,330,130 \\
\bottomrule
\end{tabular}
\end{table}
\newpage

\begin{table}[H]
\caption{Number of won, drawn, and lost positions for $7 \times 6$ \connectfour{} from the perspective of the first player. If ply is odd, then terminal positions are won otherwise lost for the first player.}
    \centering
\begin{tabular}{r|rrrrr}
\toprule
ply & won & drawn & lost & total & terminal \\
\hline
0 & 1 & 0 & 0 & 1 & 0 \\
1 & 1 & 2 & 4 & 7 & 0 \\
2 & 27 & 12 & 10 & 49 & 0 \\
3 & 35 & 58 & 145 & 238 & 0 \\
4 & 690 & 200 & 230 & 1,120 & 0 \\
5 & 1,080 & 697 & 2,486 & 4,263 & 0 \\
6 & 10,889 & 1,943 & 3,590 & 16,422 & 0 \\
7 & 17,507 & 5,944 & 31,408 & 54,859 & 728 \\
8 & 124,624 & 14,676 & 44,975 & 184,275 & 1,892 \\
9 & 197,749 & 42,896 & 317,541 & 558,186 & 19,412 \\
10 & 1,122,696 & 97,532 & 442,395 & 1,662,623 & 44,225 \\
11 & 1,734,122 & 255,780 & 2,578,781 & 4,568,683 & 273,261 \\
12 & 8,191,645 & 541,825 & 3,502,631 & 12,236,101 & 573,323 \\
13 & 12,333,735 & 1,286,746 & 17,308,630 & 30,929,111 & 2,720,636 \\
14 & 49,756,539 & 2,583,292 & 23,097,764 & 75,437,595 & 5,349,954 \\
15 & 73,263,172 & 5,596,074 & 97,682,013 & 176,541,259 & 20,975,690 \\
16 & 255,117,922 & 10,681,110 & 128,792,359 & 394,591,391 & 38,918,821 \\
17 & 369,230,362 & 21,226,658 & 467,761,723 & 858,218,743 & 130,632,515 \\
18 & 1,112,643,249 & 38,582,237 & 612,658,408 & 1,763,883,894 & 229,031,670 \\
19 & 1,589,752,959 & 70,754,712 & 1,907,752,131 & 3,568,259,802 & 670,491,437 \\
20 & 4,132,585,341 & 122,495,056 & 2,491,075,548 & 6,746,155,945 & 1,108,210,254 \\
21 & 5,849,074,428 & 208,240,707 & 6,616,029,910 & 12,673,345,045 & 2,858,601,535 \\
22 & 13,031,002,559 & 342,506,047 & 8,637,315,382 & 22,010,823,988 & 4,434,627,684 \\
23 & 18,317,405,077 & 543,074,854 & 19,402,748,258 & 38,263,228,189 & 10,130,180,393 \\
24 & 34,623,818,387 & 845,872,717 & 25,361,122,355 & 60,830,813,459 & 14,654,767,176 \\
25 & 48,376,711,901 & 1,256,717,558 & 47,632,685,500 & 97,266,114,959 & 29,672,303,474 \\
26 & 76,568,242,258 & 1,846,266,966 & 62,314,059,815 & 140,728,569,039 & 39,696,898,910 \\
27 & 106,274,173,915 & 2,578,399,088 & 96,436,935,052 & 205,289,508,055 & 71,042,927,249 \\
28 & 138,476,323,812 & 3,567,644,646 & 126,013,643,486 & 268,057,611,944 & 86,949,129,149 \\
29 & 190,301,585,678 & 4,687,144,532 & 157,638,115,456 & 352,626,845,666 & 136,563,138,602 \\
30 & 199,698,237,436 & 6,071,049,190 & 204,609,218,821 & 410,378,505,447 & 150,692,335,491 \\
31 & 269,818,663,336 & 7,481,813,611 & 201,906,000,786 & 479,206,477,733 & 205,243,451,746 \\
32 & 221,858,140,210 & 9,048,082,187 & 258,000,224,786 & 488,906,447,183 & 200,299,011,722 \\
33 & 291,549,830,422 & 10,381,952,902 & 194,705,107,378 & 496,636,890,702 & 232,494,602,432 \\
34 & 180,530,409,295 & 11,668,229,290 & 241,273,091,751 & 433,471,730,336 & 195,427,938,799 \\
35 & 226,007,657,501 & 12,225,240,861 & 132,714,989,361 & 370,947,887,723 & 188,065,840,647 \\
36 & 98,839,977,654 & 12,431,825,174 & 155,042,098,394 & 266,313,901,222 & 131,014,104,050 \\
37 & 114,359,332,473 & 11,509,102,126 & 57,747,247,782 & 183,615,682,381 & 100,184,819,358 \\
38 & 32,161,409,500 & 10,220,085,105 & 61,622,970,744 & 104,004,465,349 & 54,716,901,301 \\
39 & 33,666,235,957 & 7,792,641,079 & 13,697,133,737 & 55,156,010,773 & 31,270,711,562 \\
40 & 4,831,822,472 & 5,153,271,363 & 12,710,802,660 & 22,695,896,495 & 11,972,173,842 \\
41 & 4,282,128,782 & 2,496,557,393 & 1,033,139,763 & 7,811,825,938 & 4,282,128,782 \\
42 & 0 & 713,298,878 & 746,034,021 & 1,459,332,899 & 746,034,021 \\
\bottomrule
\end{tabular}
\end{table}

\end{document}